\documentclass[10pt,twocolumn,letterpaper]{article}

\usepackage{iccv}
\usepackage{times}
\usepackage{epsfig}
\usepackage{graphicx}
\usepackage{amsmath}
\usepackage{amssymb}
\usepackage{booktabs}
\usepackage{float}
\usepackage{enumitem}
\usepackage{multirow}
\usepackage{tabularx}
\usepackage{subcaption}

\usepackage[pagebackref=true,breaklinks=true,letterpaper=true,colorlinks,bookmarks=false]{hyperref}
\usepackage[capitalize]{cleveref}

\iccvfinalcopy 

\def\iccvPaperID{5} 

\ificcvfinal\fi

\begin{document}

\title{Using and Abusing Equivariance}

\author{Tom Edixhoven \quad Attila Lengyel \quad Jan C. van Gemert \\
Delft University of Technology}

\maketitle
\ificcvfinal\thispagestyle{empty}\fi

\begin{abstract}
    In this paper we show how Group Equivariant Convolutional Neural Networks use subsampling to learn to break equivariance to the rotation and reflection symmetries. We focus on 2D rotations and reflections and investigate the impact of the broken equivariance on network performance. 
We show that a change in the input dimension of a network as small as a single pixel can be enough for commonly used architectures to become approximately equivariant, rather than exactly.    
We investigate the impact of networks not being exactly equivariant and find that approximately equivariant networks generalise significantly worse to unseen symmetries compared to their exactly equivariant counterparts.
However, when the symmetries in the training data are not identical to the symmetries of the network, we find that approximately equivariant networks can relax their equivariance constraints, matching or outperforming exactly equivariant networks on common benchmarks.
\end{abstract}

\section{Introduction}
\label{sec:intro}
Nature contains a lot of symmetries~\cite{symmetry_nature} and networks used for computer vision have been shown to benefit greatly from prior knowledge of these symmetries. Most notably, the introduction of the convolution operator resulted in the creation of \textit{Convolutional Neural Networks} (CNN)~\cite{CNN}, form a backbone of many computer vision applications.
Convolutions are \textit{equivariant} to the translation symmetry~\cite{cnns_translation}, meaning that if an object in the input image is shifted, the output of the convolution is shifted equally. Due to translation equivariance, networks no longer have to explicitly learn to recognise objects at all possible locations, as the knowledge that location plays no role is embedded into the network.

Images, however, regularly contain other relevant symmetries for which CNNs are not equivariant. Take for example the field of histopathology, which entails the microscopic examination of organic tissue. In histopathology, the rotational orientation of the tissue is arbitrary~\cite{roto-trans_hisp}. A network that varies its output when the input is rotated is therefore a cause for uncertainty. More formally, the output of the network should be \textit{invariant} to rotation, meaning that the output should not change when the input is rotated.

\begin{figure}
  \centering
    \includegraphics[width=0.45\textwidth]{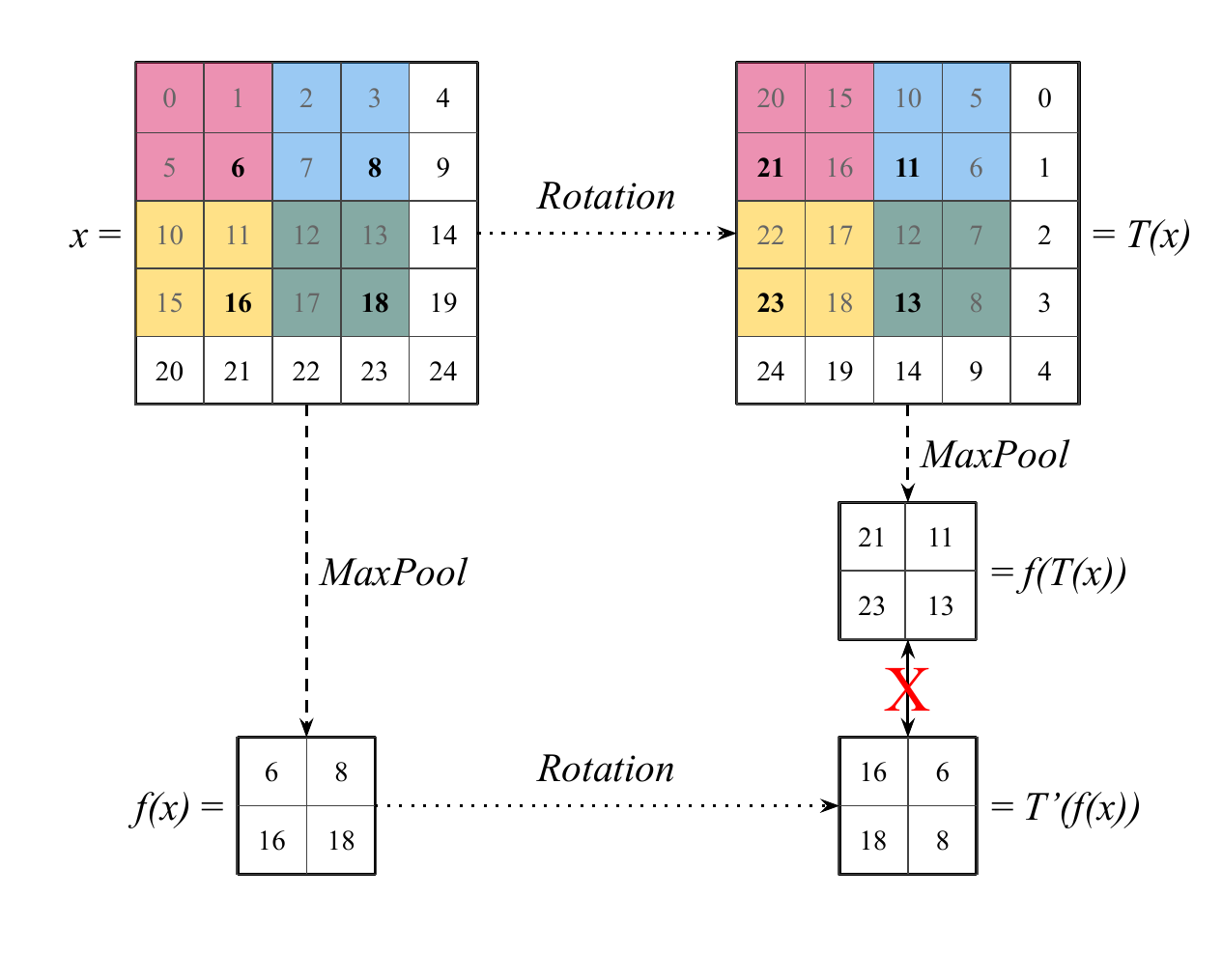}
   \caption{Example of how subsampling can break equivariance. Dotted arrows indicate a \textit{Rotation} and the dashed arrows indicate a \textit{MaxPool} subsampling layer with a kernel size and stride of 2. The locations where \textit{MaxPool} applies its pooling are coloured. One can see that $f(T(x))$ and $T'(f(x))$ contain completely different numerical values, breaking equivariance.}
   \label{fig:2d_eq_error}
\end{figure}

A major innovation in equivariance for computer vision was the introduction of \textit{Group Equivariant Convolutions} (GEC)~\cite{GCNN}, which made it possible for CNNs to guarantee equivariance or invariance to a finite group of discrete transformations, also referred to as a \textit{symmetry group}. Using GECs instead of standard convolutions to create a network yields a \textit{Group Equivariant Convolutional Neural Network} (GCNN).
Due to the group equivariant properties of GECs, GCNNs guarantee that the network output does not change when the input is rotated.

In this paper, we explore subsampling layers in GCNNs that allow the networks to break their guarantee of equivariance. Consider the \textit{MaxPool} subsampling layer in Figure~\ref{fig:2d_eq_error}. The feature map resulting from first rotating and then subsampling contains completely different numerical values than the result of first subsampling and then rotating, and as such the \textit{MaxPool} layer is not equivariant to rotations. Whether a subsampling layer breaks equivariance is dependent on the width and height of the input, also referred to as \textit{input dimension}.
Including a subsampling layer that breaks equivariance in a GCNN will void the entire GCNN's guarantee of equivariance.
However, subsampling layers are deemed almost essential for computer vision models and are used in nearly all GCNNs and modern CNNs.
Typically, no distinction is made between GCNNs that do or do not contain subsampling layers that break equivariance.
In this work, we show why a distinction should be made. We refer to networks in which subsampling layers break the guarantee of equivariance as \textit{approximately equivariant} and networks in which the guarantee is not broken are referred to as \textit{exactly equivariant}.

We offer the following contributions.
We give a formal definition of exact equivariance under subsampling and can analyse when equivariance is broken.
We show that approximately equivariant networks learn to become less equivariant  and as a result generalise significantly worse to unseen symmetries compared to their exact counterparts.
We show that slightly changing the input dimensions is often enough to make a network exactly equivariant rather than approximately equivariant.
\section{Related Work}
\subsection{Equivariance in Deep Learning}
CNNs are able to learn to become equivariant from data~\cite{data_aug, natural_equivariance}. However, this does not guarantee equivariance to the symmetries in the data and results in a redundancy in the filters of the network. For example, the network learns one filter to detect horizontal lines and a separate one to detect vertical lines, rather than a single filter to detect lines.
Much work has been written about how to efficiently teach networks equivariance to relevant symmetries during training, either by separating symmetry weights from filter weights~\cite{separate_2, separate_3, separate_1}, using contrastive learning~\cite{simCLR, equimod} or using marginal likelihood~\cite{marginal_1, marginal_2}. However, while these methods significantly increase a network's ability to become equivariant, they do not guarantee it.
Each method relies on the network learning the equivariance from the training data. However, the training data seldom guarantees a full and uniformly distributed representation of the relevant symmetries. These possible biases in the training data can then propagate into biases in the network. This can be cause for concern, as biased networks make systematic errors due to faulty assumptions about the data.

If the symmetry group for which a network needs to be equivariant is known, a common solution is to encode the symmetries into the network as prior knowledge using Group Equivariant Convolutions (GECs)~\cite{GCNN}.
GECs are equivariant to finite set of discrete transformations defined in a symmetry group.
Filter weights are then shared within the GECs according to the transformations. Because the GECs include all transformations from their symmetry group, GECs guarantee equivariance to the symmetries, regardless of biases in the training data.

The introduction of GECs kickstarted much follow-up work, including extensions from 2D planes to 3D manifolds~\cite{manifold_1, manifold_2, manifold_3} and the generalisation from discrete to continuous transformations using Lie algebra~\cite{Lie_equivariance} or other means~\cite{rotation_transformer}.
In this work, we focus on using GECs to become equivariant and invariant to the 2D roto-translation group, as the group has been proven to be useful in the field of histopathology~\cite{bekkers_roto-translation, roto-trans_hisp, PCAM_rot_eq} and processing satellite data~\cite{rot_eq_sat, deforestation}. The 2D roto-translation groups consists of all rotations and translations in a 2-dimensional space.
However, GECs are equivariant to discrete transformations and the 2D roto-translation group is continuous. Therefore, we make our networks equivariant to the \textit{p4-}group~\cite{GCNN}, consisting of all compositions of translations and $90^{\circ}$ rotations, meaning that the networks are also equivariant to rotations of $180^{\circ}$ and $270^{\circ}$.

\subsection{Breaking Equivariance}
While CNNs are generally regarded to be translation equivariant, a plethora of work has shown that this is not completely the case.
Convolutions and pooling with a stride larger than 1 have been shown to break translation equivariance~\cite{Chaman_2021_CVPR, aprox_trans_1, group_equivariant_subsampling, blurpool}. CNNs have also been shown to be able to learn absolute positions, thereby breaking the translation equivariance~\cite{osman_location}.
This is important to note, as Group Equivariant Convolutions assume that standard convolutions are translation equivariant to prove their equivariance to other transformations. 
Preventing networks from breaking their roto-translation equivariance has been investigated for reconstruction learning by introducing a group equivariant subsampling layer~\cite{group_equivariant_subsampling}. This method however requires additional compute and the effects on classification has not been investigated, where invariance is often more desirable than equivariance. In this work, we extend the current literature by investigating the influence of subsampling on roto-translation equivariance for classification.

The general proof for equivariance in GCNNs holds when the convolution convolves over the entire input. However, networks often unknowingly break this restriction. Pooling and strided convolutions are often used to aggregate local information and increase the receptive field of a network~\cite{pooling}.
The combination of stride, input size and kernel size in subsampling layers can result in different values from the input feature map being sampled, resulting in approximate equivariance rather than exact equivariance~\cite{romero2020attentive}.
While it might seem like a minute detail, we find that it causes GCNNs to underperform relative to other equivariant networks in related works.
Examples of rotation equivariant GCNNs exhibiting unexpected behaviour can be found in~\cite{c3po, RIC-CNN}.
In this work, we show that we can guarantee equivariance for GCNNs by introducing a relatively simple restriction on the combination of input size, kernel size and stride.

\subsection{Relaxing Equivariant Constraints}
Recent work has shown the possible benefits of relaxing equivariant constraints, showing that networks can gain performance by allowing them to learn to become less equivariant~\cite{romero_partial_eq, relaxing_eq}.
This is relevant for our work, since being approximately equivariant seems to allow networks to relax their own equivariant constraints. Similarly to the aforementioned works, our solution also enables the user to make a conscious decision whether to relax the equivariant constraints on a network.

\section{How subsampling breaks equivariance}
In this section we provide a more formal introduction to equivariance and Group Equivariant Convolutions (GConvs). We will then show a simple network configuration including a subsampling operation that breaks the equivariance property of GConvs. Subsequently we introduce a constraint on the network configuration that does guarantee exact equivariance under subsampling, and provide a proof for rotations and mirroring.

\subsection{Group Equivariant Convolutions}\label{ssec:measuring_equivariance}
A network $f$ is equivariant to transformation $T$, when the output of $f$ on input $x$ changes predictably when $x$ is transformed by $T$. More formally, there exists a transformation $T'$ for which the following equality holds:
\begin{equation}
    f(T(x)) = T'(f(x)).
    \label{eq:equivariance}
\end{equation}

GCNNs are equivariant to a set of transformations defined in a symmetry group $G$, where in practice the transformations are stored in an additional group dimension in the feature maps.
In the case of the \textit{p4-}group, $T'$ consists of a rotation in the spatial dimensions and permutation of the group dimension on the feature map. Invariance to $T$ is achieved by applying a coset pooling operation on the group dimension, such that the final representation satisfies
\begin{equation}
    f(T(x)) = f(x).
    \label{eq:invariance}
\end{equation}
As such, invariance is considered a special case of equivariance.

\subsection{Exact and Inexact Equivariance}\label{ssec:achieving_eq}
From Equation~\ref{eq:equivariance} and the cyclic permutation defined by $T'$ it follows that the feature maps of an exactly equivariant GCNN should contain the same numerical values regardless of the applied rotation $T$ to the input.
However, layers that perform subsampling on the input can introduce numerical differences depending on the input transformation, resulting in the network architecture no longer being exactly equivariant, as we will now demonstrate.

Let $x$ be a rectangular input of dimension $i=5$, $f$ a network consisting of a single \textit{MaxPool} layer with kernel size $k=2$ and stride $s=2$, and $T$ a clockwise rotation of $90^{\circ}$. As the reference point of the 2D subsampling operation is always defined at the $(0,0)$ index of the input, applying $T$ results in the sampling indices being shifted by a single pixel from the perspective of the \textit{MaxPool} layer. Subsequently, $T'(f(x))$ and $f(T(x))$ contain different numerical values, as shown in Figure~\ref{fig:2d_eq_error}. To guarantee exact equivariance we therefore need to ensure that the same indices are sampled, irrespective of the order in which the sampling operation and rotation are applied.
Our proposed solution is simple as it does not require any modifications to the network architecture and relies purely on setting appropriate input dimensions to the network.
For comprehensibility, we focus on the case of square inputs $\in \mathbb{R}^{i \times i}$ and an arbitrary kernel size $k$ and stride $s$, but the proof can be readily extended to rectangular inputs $\in \mathbb{R}^{j \times i}$.

A GCNN is exactly equivariant to rotations of multiples of $90^{\circ}$ if the following equation holds for all layers in the network:
\begin{equation}
  (i - k)\mod s = 0.
  \label{eq:exactly_equivariant}
\end{equation}


We prove that Equation~\ref{eq:exactly_equivariant} is a necessary condition for exact equivariance to $90^{\circ}$ rotations by asserting that the sampled indices for a given output index remain the same under rotation. We define a new function called index, that returns the indices of the input values used by a convolutional or pooling layer to calculate the value located at index $(x, y)$ in the output:
\begin{equation}
    \label{eq:sampled_indices_compact}
    \text{index}\left(\begin{bmatrix} x \\ y \end{bmatrix}\right) = \left[
        \begin{bmatrix} sx \\ sy\end{bmatrix},  \begin{bmatrix} sx+k-1 \\ sy+k-1\end{bmatrix}\right].
\end{equation}
Here $s$ is the stride used for subsampling and $k$ represents the kernel size. The output of the function is a square patch, denoted as $[\vec{u}, \vec{v}]$, where $\vec{u}$ and $\vec{v}$ represent the indices of the top left and bottom right corner, respectively. The sampled indices include all integer tuples within this patch.
We also introduce the function $R$, which takes an index $(x,y)$ as input and returns the indices rotated $90^{\circ}$ counterclockwise:
\begin{equation}
\label{eq:r90_index}
    R_n\left(\begin{bmatrix} x \\ y \end{bmatrix}\right)=\begin{bmatrix}
           y \\
           n-1-x
         \end{bmatrix},
\end{equation}
where $n$ indicates the width and height of the feature map in which the index $(x, y)$ is located.
We further generalise Equation~\ref{eq:r90_index} to an input patch $[\vec{u}, \vec{v}]$ rather than a single coordinate, resulting in Equation~\ref{eq:r90_indices}:
\begin{align}
    \label{eq:r90_indices}
    \begin{split}
    R_n\left(\left[\vec{u}, \vec{v}\right]\right) &= R_n\left(\left[\begin{bmatrix} x_1 \\ y_1 \end{bmatrix}, \begin{bmatrix} x_2 \\ y_2 \end{bmatrix}\right]\right) \\
    &= \left[\begin{bmatrix} y_1 \\ n-1-x_2 \end{bmatrix}, \begin{bmatrix} y_2 \\ n-1-x_1 \end{bmatrix}\right]
    \end{split}
\end{align}
In the resulting output coordinates $x_1$ and $x_2$ get interchanged due to the counterclockwise rotation of the patch: the top left corner becomes the bottom left corner, while the bottom right corner becomes the top right corner.

Given that our layer takes a feature map with a width and height of $i$ as input, we can write the width and height of the output feature map as
\begin{equation}
    o=\lfloor \frac{i-k}{s}\rfloor + 1.
\end{equation}

For a layer to be exactly equivariant, determining the sampled indices and then rotating should return the same result as rotating first and then determining the sampled indices, which we can formally denote as
\begin{equation}
\label{eq:subsampling_rot_eq}
    \text{index}\left(R_o\left(\begin{bmatrix} x \\ y \end{bmatrix}\right)\right) = R_i\left(\text{index}\left(\begin{bmatrix} x \\ y \end{bmatrix}\right)\right).
\end{equation}
To solve the left-hand side, we substitute Equation~\ref{eq:r90_index} into Equation~\ref{eq:sampled_indices_compact}, yielding
\begin{align}
    \label{eq:expand_first_term}
    \begin{split}
    \text{index}\left(R_o\left(\begin{bmatrix} x \\ y \end{bmatrix}\right)\right) = \\
    \text{index}\left(\begin{bmatrix} y \\ \lfloor \frac{i-k}{s}\rfloor - x \end{bmatrix}\right) = \\
    \left[ \begin{bmatrix} sy \\ s\lfloor \frac{i-k}{s}\rfloor -sx \end{bmatrix}, \begin{bmatrix} sy+k-1 \\s\lfloor \frac{i-k}{s}\rfloor -sx+k-1 \end{bmatrix} \right].
    \end{split}
\end{align}
The same can be done for the right-hand side, by substituting Equation~\ref{eq:sampled_indices_compact} into Equation~\ref{eq:r90_indices}, resulting in
\begin{align}
    \label{eq:expand_second_term}
    \begin{split}
        R_i\left(\text{index}\left(\begin{bmatrix} x \\ y \end{bmatrix}\right)\right) = \\
        R_i\left(\left[\begin{bmatrix} sx \\ sy \end{bmatrix}, \begin{bmatrix} sx+k-1 \\ sy+k-1 \end{bmatrix}\right]\right) = \\
        \left[\begin{bmatrix} sy \\ i-k-sx \end{bmatrix}, \begin{bmatrix} sy+k-1 \\ i-1-sx \end{bmatrix}\right].
    \end{split}
\end{align}

Substituting Equations~\ref{eq:expand_first_term} and~\ref{eq:expand_second_term} into Equation~\ref{eq:subsampling_rot_eq}, we find two equations
\begin{equation}
    s\lfloor \frac{i-k}{s}\rfloor - sx = i - k - sx,
\end{equation}
\begin{equation}
    s\lfloor \frac{i-k}{s}\rfloor - sx + k - 1 = i - 1 - sx.
\end{equation}
Removing duplicate terms yields a single equation
\begin{equation}
\label{eq:subsampling_rot_eq_conclusion}
    s\lfloor \frac{i-k}{s}\rfloor = i-k,
\end{equation}
which can be simplified to Equation~\ref{eq:exactly_equivariant}. As Equation~\ref{eq:exactly_equivariant} holds for a rotation of $90^{\circ}$, it automatically holds for rotations of $180^{\circ}$ and $270^{\circ}$, as these can be composed using multiple rotations of $90^{\circ}$. We provide a similar proof for mirroring in the supplementary material.

\subsection{Measuring Equivariance and Invariance}\label{ssec:measuring_final_invariance}
We evaluate the exactness of the equivariance property of GConvs both in terms of their measured invariance and equivariance.

\paragraph{Measuring equivariance}
Since in GConvs both $T$ and $T'$ are known transformations, feature maps $f(T(x))$ and $T'(f(x))$ can be computed independently. We can thus define the equivariance error in terms of the Mean Squared Error (MSE) between the two feature maps:
\begin{align}
\label{eq:frobenius}
\begin{split}
\epsilon &= \text{MSE}(f(T(x)), T'(f(x))) \\
&= \frac{1}{ijk} \sqrt{\sum_{i, j, k}|f(T(x))_{ijk} - T'(f(x))_{ijk}|^2},
\end{split}
\end{align}
%
where $i$ and $j$ sum over the spatial dimensions of the feature map and $k$ sums over the group dimension. The equivariance error can be evaluated at any GConv layer in the network.

\begin{figure}
  \centering
   \includegraphics[width=0.2\textwidth]{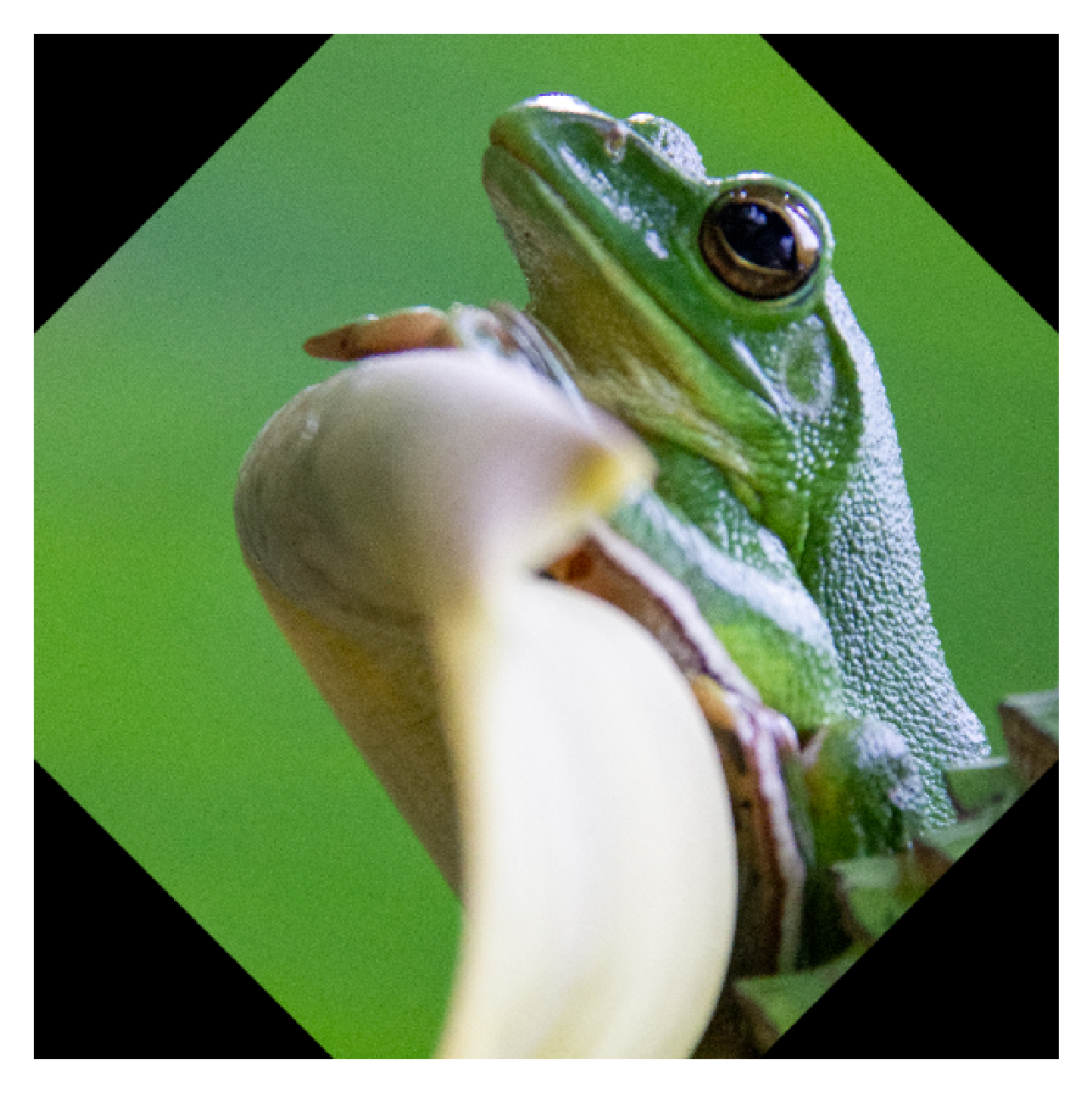}
    \includegraphics[width=0.2\textwidth]{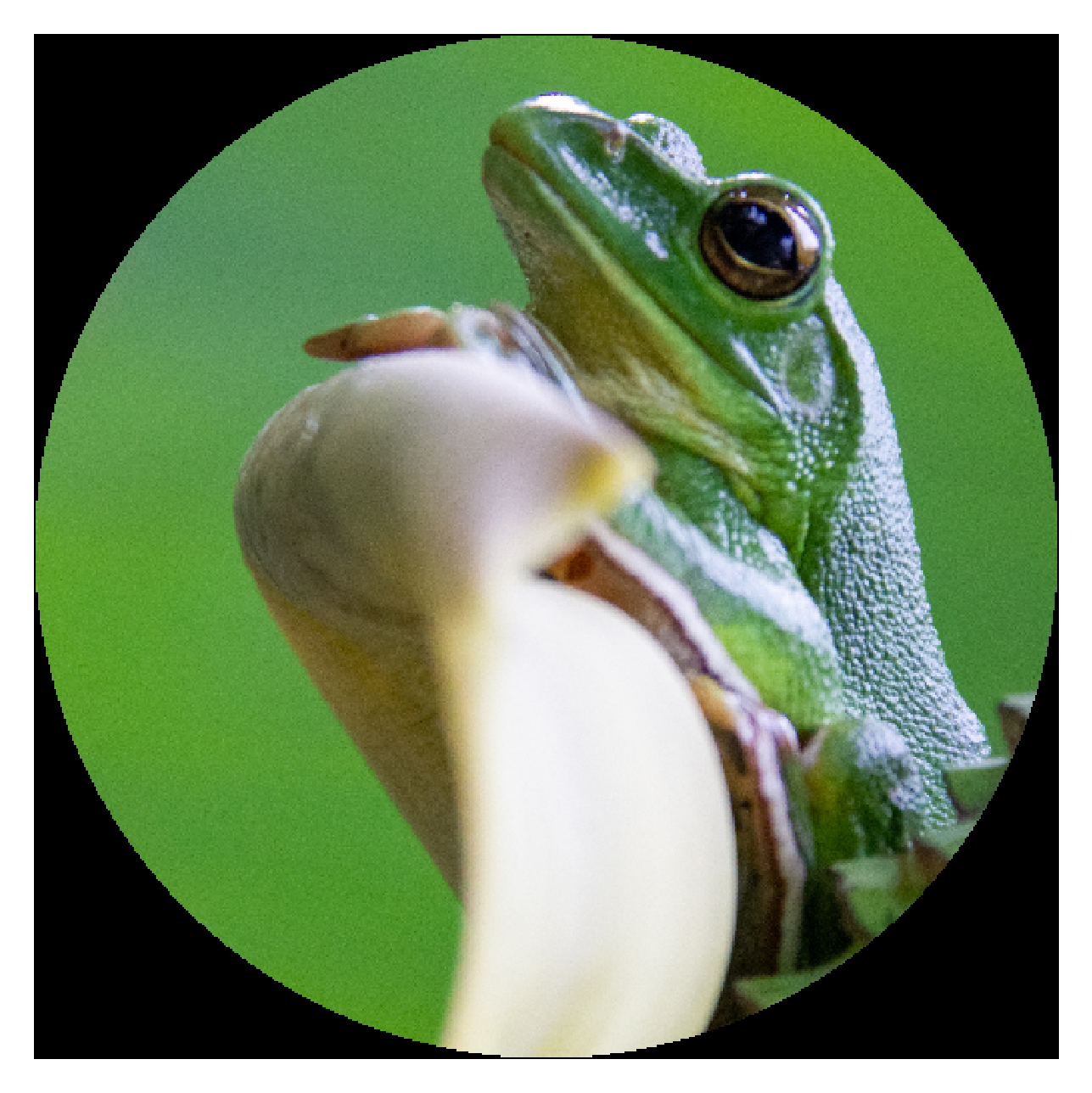}
   \caption{\textbf{Left:} Rotated input without \textit{CircleCrop}. \textbf{Right:} Rotated input with \textit{CircleCrop}. Without \textit{CircleCrop} the rotation is distinctive.}
   \label{fig:circle_crop}
\end{figure}

\paragraph{Measuring invariance}
To measure the invariance of the network output after coset pooling we apply a range of rotations in $[0^{\circ}, 360^{\circ})$ to the test set and report the test accuracy for each set separately. However, rotating an image by degrees other than multiples of $90^{\circ}$ introduces artefacts at the corners of the image, as shown in Figure~\ref{fig:circle_crop} (left). These artifacts have an additional detrimental effect on the network's performance. To ensure we only measure the performance drop due to rotation, we apply a \textit{CircleCrop}, which sets all values whose coordinates are not inside the largest possible inscribed circle to 0, as visualised in Figure~\ref{fig:circle_crop} (right). To prevent any domain shift between the train and test set, we apply \textit{CircleCrop} during both training and evaluation. \textit{Nearest Neighbour Interpolation} also affected model performance and so all rotations are performed using \textit{Bilinear Interpolation}.

\section{Experiments}
\subsection{Breaking Equivariance}\label{ssec:breaking_eq}
In this subsection, we show how networks can learn to break their equivariance to improve their performance, and demonstrate that they in practice do so on commonly used classification datasets.

\paragraph{Can GCNNs break equivariance?}
If a GCNN is truly invariant, it should be unable to distinguish between an input $x$ and $T(x)$, where $T$ is a $90^{\circ}$ rotation. We challenge this assumption by explicitly training a simple GCNN to differentiate between the two input samples $x$ and $T(x)$. We construct a network consisting of
(i) a GConv layer with $k=3$, $s=2$, 1 output channel and a padding of 1; (ii) a global average pooling layer over the spatial dimensions; (iii) a coset max pooling layer over the group dimension to obtain a rotation invariant representation; and (iv) a fully connected layer with two output features.



\begin{figure}
  \centering
    \includegraphics[width=0.4\textwidth]{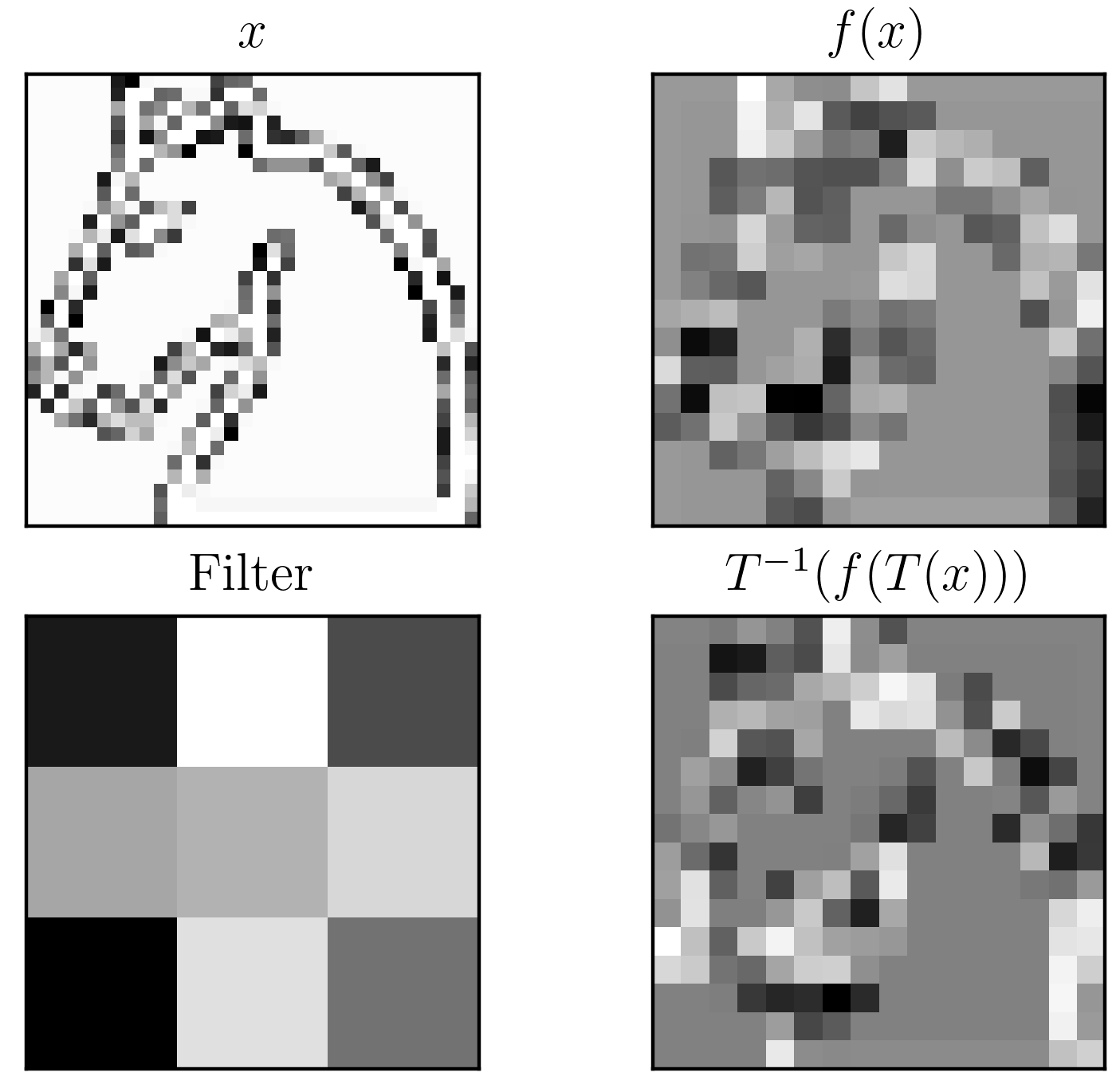}
   \caption{A subsampling Group Equivariant net $f$ that is equivariant to the $90^{\circ}$ rotation transformation $T$ can learn a filter that returns almost inverted values for $f(x)$ and $T^{-1}(f(T(x)))$, while these outputs should be identical in theory. Because the outputs are not equal, a network using the filter $f$ can perfectly distinguish between $x$ and $T(x)$.}
   \label{fig:osman_rot}
\end{figure}

We define an input $x_1 \in \mathbb{R}^{32\times32}$ as shown in the top left of Figure~\ref{fig:osman_rot} and train the network on $x_1$ and $T(x_1)$. We find that the network can perfectly distinguish between the two samples as shown by the feature maps in the right column in Figure~\ref{fig:osman_rot}. This demonstrates that the network is not invariant, despite the pooling operation on the group dimension. We furthermore define a second input $x_2 \in \mathbb{R}^{33\times33}$ and repeat the experiment. The network is not able to distinguish between $x_2$ and $T(x_2)$, showing that the same network architecture is exactly invariant for inputs in $\mathbb{R}^{33\times33}$, while not being invariant for inputs in $\mathbb{R}^{32\times32}$. 
This is in line with our findings in Section~\ref{ssec:achieving_eq}, as for $k=3$, $s=2$ and $i=33$
\begin{align*}
    (i - k)\mod s = (33 - 3)\mod 2 = 0
\end{align*}
holds, whereas for $i=32$
\begin{align*}
    (i - k)\mod s = (32 - 3)\mod 2 = 1 \neq 0
\end{align*}
does not. Thus, GCNNs can break equivariance.

\paragraph{Breaking equivariance on common benchmarks.}
We have shown that when the objective of a GCNN is to break its equivariance, it will do so if possible. However, the question remains whether a network will also learn to do so when breaking equivariance is not explicitly the objective.

\begin{figure*}
	\centering
	\begin{subfigure}{0.5\textwidth}
		\centering
		\includegraphics[width=0.8\linewidth]{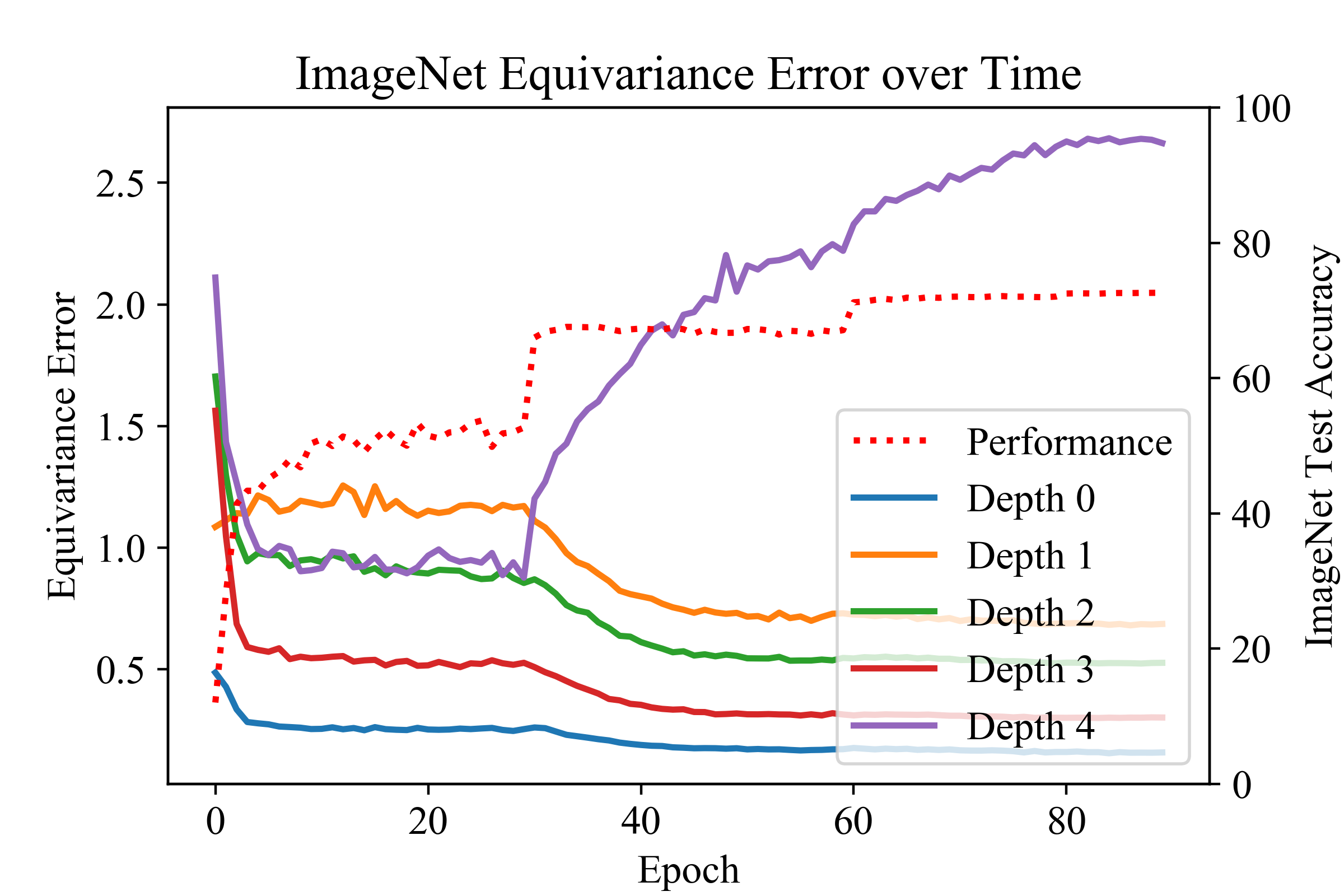}
		\caption{}
        \label{fig:eq_er_time_ImageNet}
    \end{subfigure}%
	\begin{subfigure}{0.5\textwidth}
		\centering
        \includegraphics[width=0.8\linewidth]{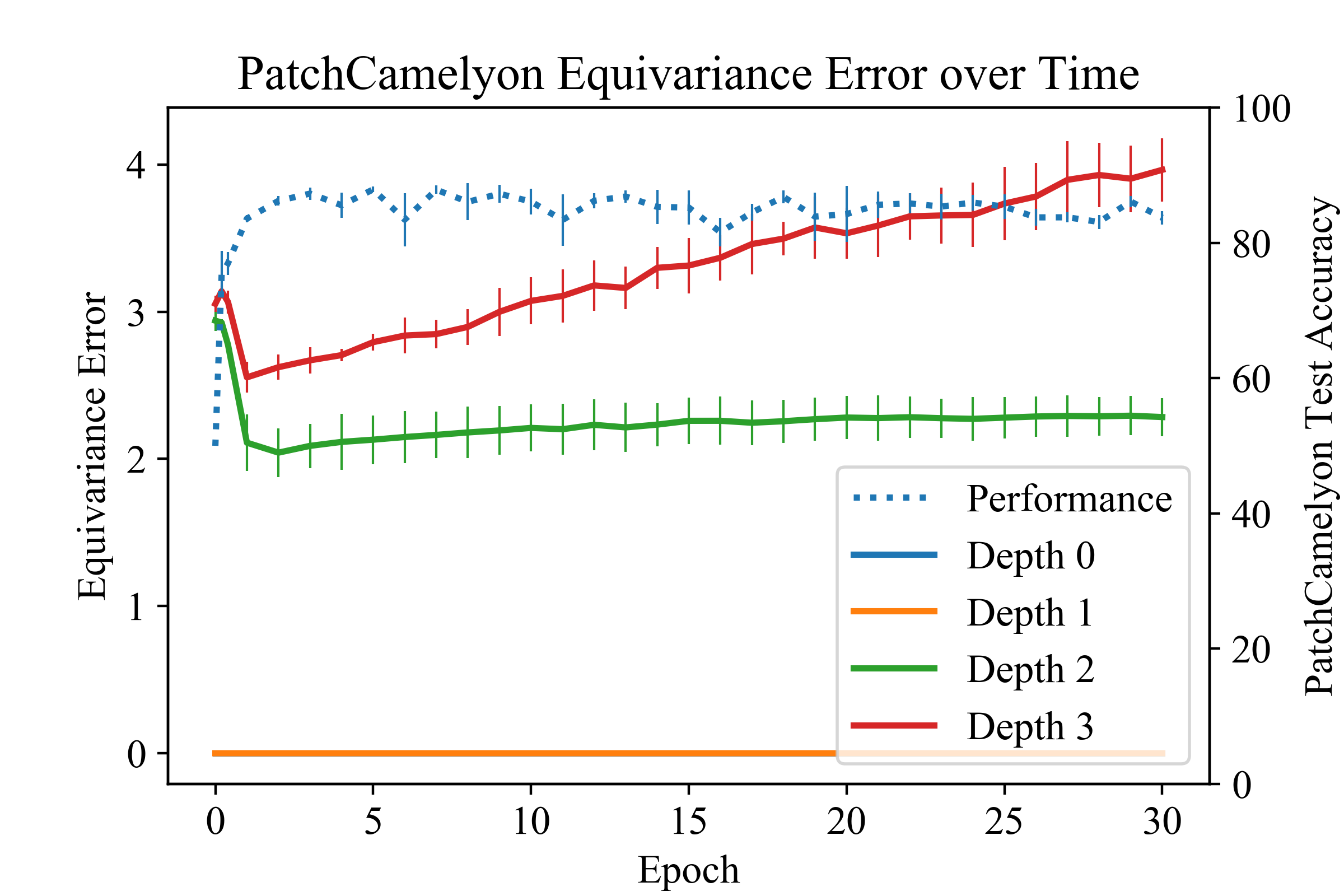}
        \caption{}
        \label{fig:eq_er_time_PCAM}
    \end{subfigure}
    \caption{The measured equivariance error at different depths in a $p4$-ResNet, trained on ImageNet (a) and PatchCamelyon (b), respectively. The classification accuracy is indicated as a dotted red line. The equivariance error in the final layer increases throughout training, indicating that the network is learning to become less equivariant, even when trained on the rotation invariant PatchCamelyon dataset.}%
\end{figure*}

We first investigate the ImageNet~\cite{imagenet} classification problem. We create a rotation equivariant ResNet18~\cite{resnets} by substituting standard convolutions with \textit{p4-}convolutions. The network width is divided by $\sqrt{4}$ to keep the number of parameters approximately equal to a standard ResNet18 and the input images are kept at their original $224 \times 224$ size. Training is performed for 90 epochs using the default training settings, i.e. SGD with momentum 0.9 and learning rate 0.1, which is step-wise reduced by a factor 0.1 every 30 epochs. The network is trained to classify the standard ImageNet classes, so there is no explicit objective to distinguish between rotations. Throughout training we monitor the equivariance error as defined in Equation~\ref{eq:frobenius} after the first layer and each of the four ResNet stages.
The measured equivariance errors are shown in Figure~\ref{fig:eq_er_time_ImageNet}. We observe that initially all equivariance errors drop to a more or less constant value, and upon decreasing the learning rate at epoch 30 the equivariance error further drops in most stages in the network. However, the error at the last stage of the network increases rapidly, finally plateauing at a value higher than after random initialization.

Secondly, we look at the PatchCamelyon dataset~\cite{PCAM_rot_eq}. This dataset is interesting because it is pathology data, which, unlike ImageNet images, should not contain any dominant rotation bias.
We use a similar setup to our previous ImageNet experiment, but we replace the ResNet18 with a ResNet44. The network width is decreased to obtain an approximately equal number of parameters as in the architecture used by Veeling \etal~\cite{PCAM_rot_eq} when evaluating on PatchCamelyon. The results on PatchCamelyon can be found in Figure~\ref{fig:eq_er_time_PCAM}. As the first layer and the first stage both have a stride of 1, the equivariance error is a constant 0 at the first two depth measurements. The other two stages show a similar behavior as in the ImageNet experiment, with a rapid decrease in the initial part of the training and a gradual decrease in the final stage throughout the remainder of the training run. Interestingly, even though PatchCamelyon is a rotation invariant dataset by definition, the network still learns to break its equivariance. 

We can thus conclude that a network that is equivariant to rotations can learn to abuse its approximate equivariance to become less equivariant to rotations. This occurs both when the network is trained on ImageNet, a dataset which contains only a limited amount of rotations and a clear upright orientation bias, as well as the PatchCamelyon dataset which should be rotation invariant by definition. A network learning differences between rotations in a rotation invariant setting is cause for concern, as the rotations are arbitrary and therefore should not contain any relevant information.
 

\subsection{Impact of Exact Equivariance}

\paragraph{Performance on unseen rotations}\label{sssec:ood_perf} 
Due to the discrete nature of GCNNs, it is impossible to include all continuous rotations in the discrete group. Thus, it is important to generalise well to rotations that are not part of the group dimension. To compare how well approximately and exactly equivariant networks generalise to unseen rotations, we perform a controlled experiment on the MNIST~\cite{mnist} dataset of handwritten digits. 
Since MNIST contains limited rotations due to slanted handwriting, we are able to control what rotations are included during training and testing by transforming the data.

For this experiment, we use the Z2CNN and P4CNN architectures introduced by Cohen \etal~\cite{GCNN}. The Z2CNN consists of 6 layers of $3 \times 3$ convolutions, followed by a single $4 \times 4$ convolutional layer, each layer consisting of 20 channels. Each layer is followed by a ReLU activation and batch normalisation layer. A dropout layer with $p=0.3$ is added after layers 1 through 5, and a max-pooling layer with a stride of 2 after the second layer. The convolutional part is followed by a global spatial average-pooling layer, and lastly, a fully connected layer. The P4CNN architecture is created by substituting standard convolutions with \textit{p4}-convolutions and introducing a group coset max-pooling layer before the fully connected layer. To keep the number of parameters of Z2CNN and P4CNN approximately equal, the number of channels in P4CNN is divided by $\sqrt{4}$. We use the default input size of $28\times28$ for exact equivariance, and input sizes $27\times27$ and $29\times29$ for approximate equivariance. The results are averaged over 10 runs with different random seeds. The models are trained for 50 epochs using Adam~\cite{adam} and an initial learning rate of 0.01, which is halved every 10 epochs.

\begin{figure}
  \centering
    \includegraphics[width=0.4\textwidth]{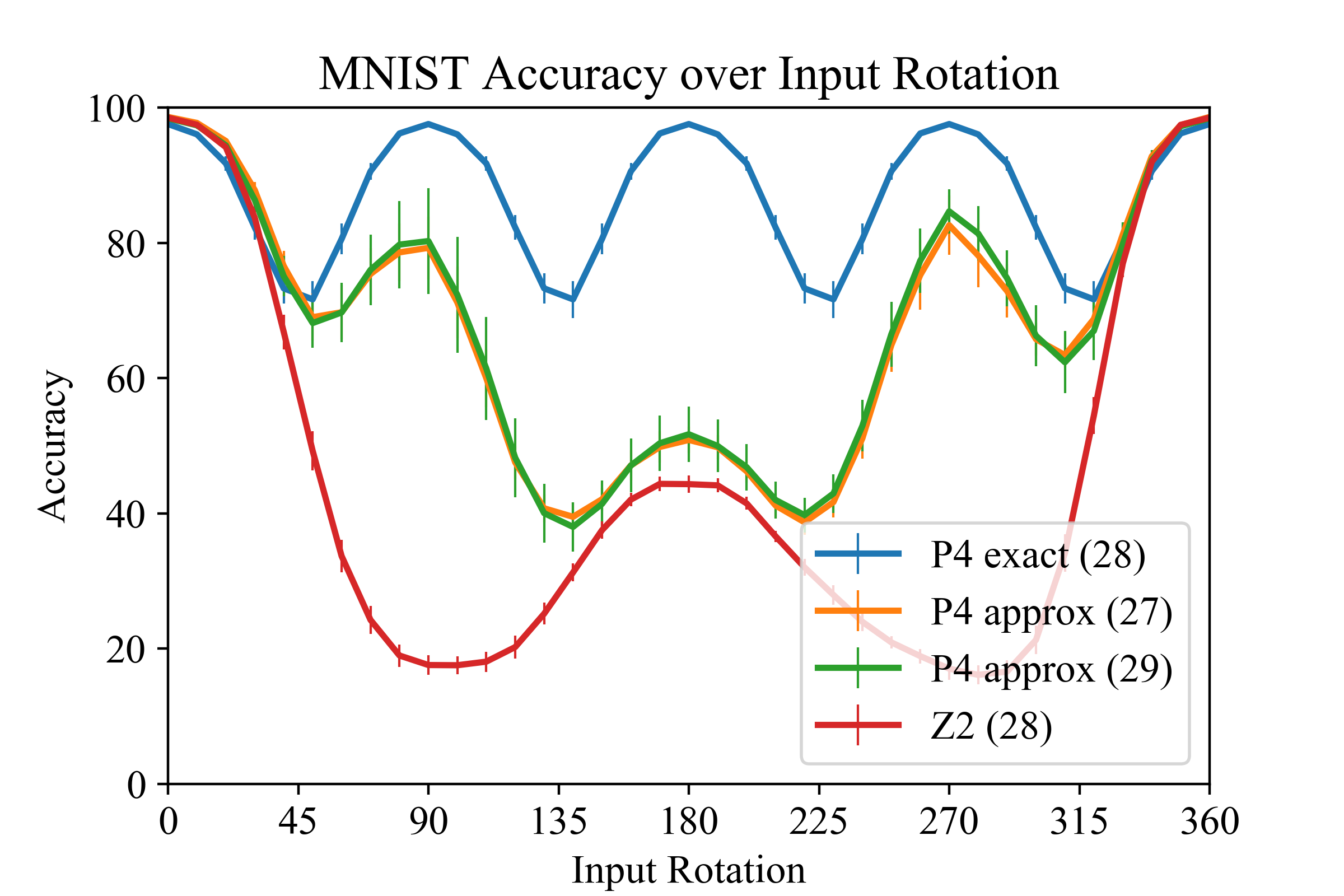}
   \caption{The equivariance of a network can be evaluated by explicitly applying the transformation on the test set.
   When the training contains no data augmentations, an exactly equivariant network generalises significantly better than its approximate counterpart.}
   \label{fig:non_rot_training}
\end{figure}

The results in Figure~\ref{fig:non_rot_training} show the performance of a model trained on MNIST and evaluating on RotMNIST, a uniformly rotated version of MNIST. An exactly equivariant network will significantly outperform approximate counterparts on rotated samples. All the \textit{p4-}equivariant networks still outperform the Z2CNN baseline. We also observe a much higher standard deviation in the performance of the approximately equivariant networks. The performance increase of Z2CNN at $180^{\circ}$ can be attributed to the rotational symmetries in the MNIST dataset. The $0, 1$ and $8$ classes stay roughly identical when rotated $180^{\circ}$.

\begin{figure*}
  \centering
  \centering
   \includegraphics[width=0.4\textwidth]{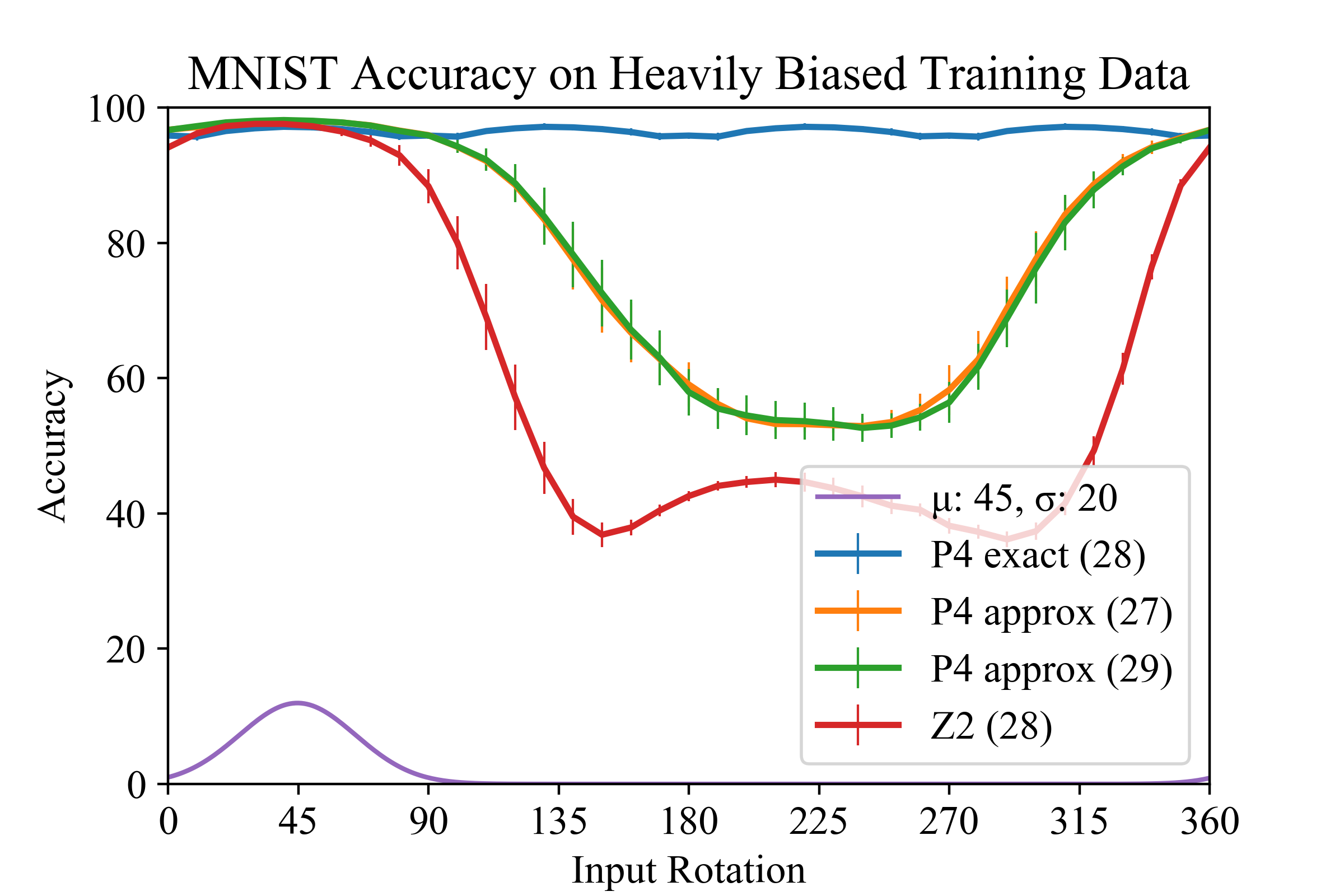}
    \includegraphics[width=0.4\textwidth]{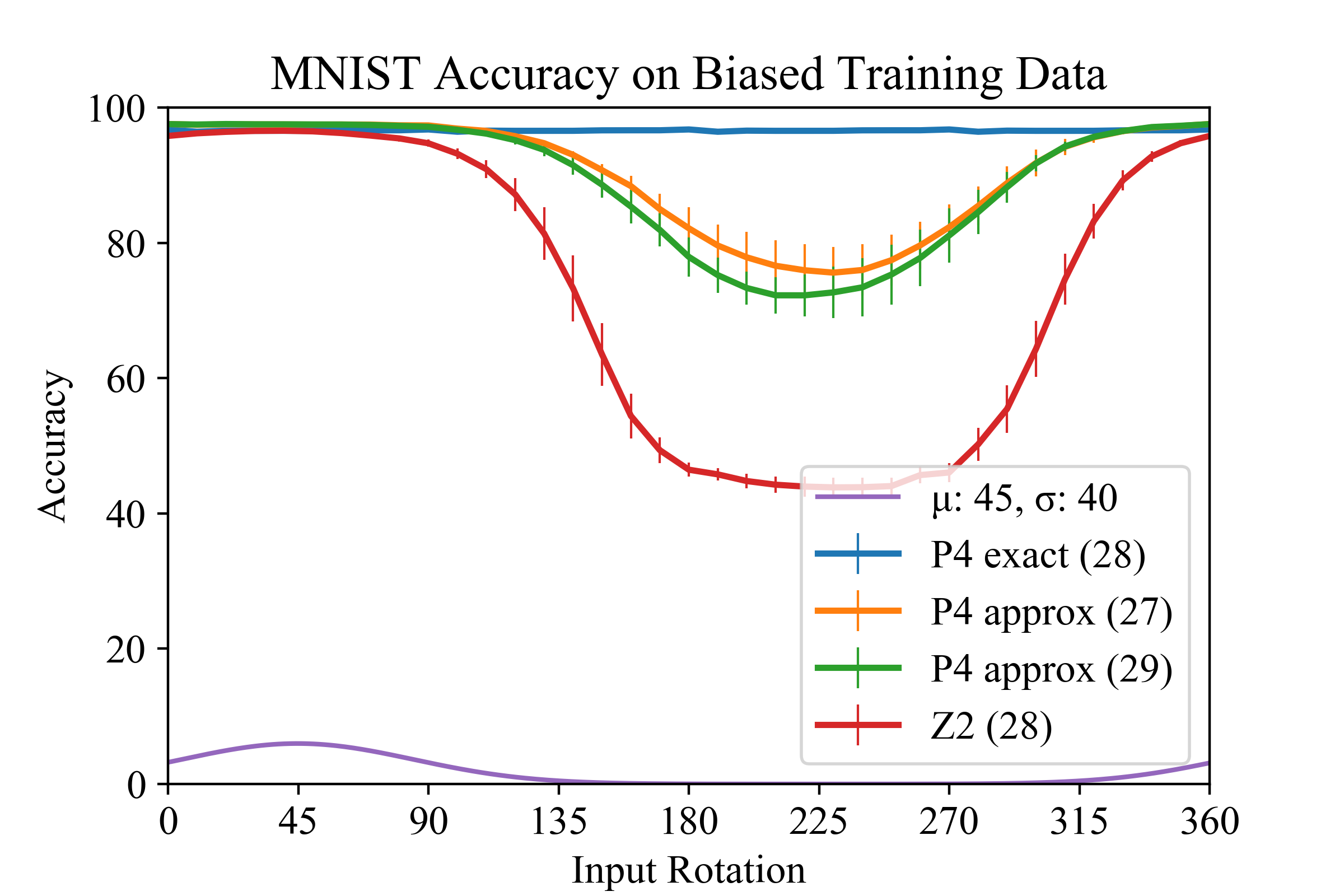}
   \caption{An exactly equivariant network generalises significantly better to unseen rotations in case of a rotation bias in the training data.
   The rotation angles of the training set are sampled from a normal distribution $\mathcal{N}(\mu, \sigma)$, visualised at the bottom of each plot. 
   The plot on the left shows the performance on a biased distribution with $\sigma=20$. The plot the right shows the performance on a biased distribution with $\sigma=40$, which is less severe but still significant.
   }
   \label{fig:biased_rot_training}
\end{figure*}

To further evaluate network generalisability to unseen rotations, we create two new versions of RotMNIST with biased rotation transformations. The rotation of each training digit is sampled from a normal distribution. Both datasets use a mean rotation of $45^{\circ}$, one has a standard deviation of $20^{\circ}$ and the other of $40^{\circ}$. We then train the networks on these biased training sets and evaluate them on a test set with uniform rotations.  The results in Figure~\ref{fig:biased_rot_training} show that, similarly to training on non-rotated data, the exactly equivariant network generalises noticeably better than the others. The exactly equivariant network almost becomes invariant to rotations in general, while all other networks exhibit a significant drop in performance on rotations that are not in the training data. Since the transformation distributions in the training data are often unknown, it is important to generalise to instances of the transformation that are not in the training data.

We further evaluate the impact of exact and approximate equivariance on common benchmark datasets, where we differentiate between datasets with and without rotational symmetries. Datasets with rotations include \textit{Flowers102}~\cite{flowers102}, where many classes have a rotationally symmetric shape, and \textit{PatchCamelyon}~\cite{PCAM_rot_eq}, which is completely invariant to rotation. Datasets without rotational symmetries include \textit{Cifar10}, \textit{Cifar100}~\cite{cifar} and \textit{ImageNet}~\cite{imagenet}.
We evaluate for unseen rotations by rotating the test set by multiples of $90^{\circ}$ and averaging the performance over the four rotations. To achieve exact and approximate p4-equivariance, we change the input size of the network, such that equation 3 holds for all layers in the network. The results are shown in Table~\ref{tab:natural_data_application}. The approximately equivariant networks are outperformed by the exactly equivariant networks in all cases. The difference is most significant for datasets that contain limited rotations, i.e. \textit{Cifar10}, \textit{Cifar100} and \textit{ImageNet}, as here the model is not able to learn the symmetries from data. Both the exact and inexact networks outperform the baseline CNN by a large margin, showing that in general inexact equivariance is a beneficial property.

\paragraph{Performance on seen rotations}
For the performance on rotations that are included in the training data, also referred to as seen rotations, we first evaluate MNIST and RotMNIST using P4CNN and Z2CNN as in Section~\ref{sssec:ood_perf}. We report the mean and standard deviations of the test accuracies over 100 run with different random seeds in Table~\ref{tab:exact_impact_MNIST}. On MNIST the exactly equivariant network exhibits a performance drop between $0.65\%$  and $0.91\%$ compared to its approximately equivariant counterparts, which is confirmed to be statistically significant, as shown in the supplementary material. On RotMNIST the exact network performs identically to the approximate networks, as the approximate networks are able to learn to become invariant from the transformations found in the training data.

\begin{table}
  \centering
  \begin{tabularx}{\linewidth}{@{}lXXX@{}}
    \toprule
    Model & Equivar. & MNIST & RotMNIST \\
    \midrule
    Z2CNN & - (28) & $98.47 \pm 0.17$ & $91.60 \pm 1.25$ \\
    P4CNN & Approx (27) &  $98.52 \pm 0.26$  & $96.92 \pm 0.27$ \\
    P4CNN & Exact (28) & $97.69 \pm 0.17$ & $96.89 \pm 0.21$ \\
    P4CNN & Approx (29) &  $98.42 \pm 0.25$ & $96.87 \pm 0.25$ \\
    \bottomrule
  \end{tabularx}
  \caption{Network accuracy denoted as mean $\pm$ standard deviation on MNIST and RotMNIST test sets. The standard deviation is calculated using a 100 runs with different seeds. The equivariance column indicates whether the network is exactly or approximately equivariant and contains the network input size in parentheses.}
  \label{tab:exact_impact_MNIST}
\end{table}

To evaluate for seen rotations on common classification benchmarks we compute the model accuracy on the default test set. The benchmark results can be found in Table~\ref{tab:natural_data_application}. The exactly equivariant networks are generally matched or outperformed by their approximately equivariant counterparts, even on datasets containing rotational symmetries. This seems to indicate that there lies value in relaxing the equivariant constraints of networks. Furthermore, both \textit{p4}-equivariant networks outperform their \textit{z2}-equivariant counterparts, even when the dataset is not known for containing many rotational symmetries. This could indicate that the improvements from the group equivariant architecture might not be solely from equivariance, but could also originate from other traits of GCNNs. Other possible explanations are the increase in computations or the amount of gradients a GCNN uses compared to a standard CNN.

\begin{table*}
  \centering
  \begin{tabularx}{\linewidth}{@{}llXXX|XXX@{}}
    \toprule
  \textbf{Dataset} & \textbf{Model} &
  \multicolumn{3}{c}{\textbf{Unseen rotations}} &
  \multicolumn{3}{c}{\textbf{Seen rotations}} \\
  & & Approx. $p4$ & Exact $p4$ & Standard & Approx. $p4$ & Exact $p4$ & Standard \\ \midrule
    Flowers102 & ResNet-18 & 83.32 $\pm$ 1.21 & \textbf{86.65 $\pm$ 1.41} & 68.78 $\pm$ 0.29 & 86.28 $\pm$ 1.32 & \textbf{86.65 $\pm$ 1.41} & 82.18 $\pm$ 0.53 \\
  PCam & ResNet-44 & 86.66 $\pm$ 0.72 & \textbf{87.40 $\pm$ 0.71} & 82.43 $\pm$ 1.71 &  \textbf{87.52 $\pm$ 1.20} & 87.40 $\pm$ 0.71 & 85.35 $\pm$ 1.04 \\
  CIFAR10 & ResNet-44 & 79.19 $\pm$ 0.47 & \textbf{93.41 $\pm$ 0.09} & 52.68 $\pm$ 0.11 & \textbf{94.80 $\pm$ 0.21} & 93.41 $\pm$ 0.09 & 93.20 $\pm$ 0.11 \\
  CIFAR100 & ResNet-44 & 58.82 $\pm$ 0.36 & \textbf{72.28 $\pm$ 0.40} & 38.62 $\pm$ 0.16 & \textbf{75.00 $\pm$ 0.52} & 72.28 $\pm$ 0.40 & 70.09 $\pm$ 0.28 \\
  ImageNet & ResNet-18 & 60.01 & \textbf{72.55} & 48.10 & 72.48 & \textbf{72.55} & 70.00 \\
    
  \end{tabularx}
  \caption{Test accuracies of ResNet models on seen and unseen rotations of common classification benchmarks. Exactly equivariant networks perform significantly better on unseen rotations, especially on datasets containing no rotation symmetries, i.e. CIFAR and ImageNet. Approximate networks are able to relaxe equivariance constraints and perform better on seen rotations. All \textit{p4-}equivariant networks outperform the standard CNN.}
  \label{tab:natural_data_application}
\end{table*}

\section{Conclusion}
In this work, we show that Group Equivariant Convolutions~\cite{GCNN} can and do learn to break their equivariance towards the 2D rotations in common use cases.
We prove theoretically and empirically that changing the input size of the network is sufficient to prevent a network from breaking its equivariance.
We find that exactly equivariant networks generalise significantly better to unseen rotations than their approximately equivariant counterparts, but that when the training data contains all relevant rotations there is no significant difference. We find that due to broken translation equivariance, the rotation equivariance of GCNNs is also broken. It could prove interesting to further investigate the effect of making networks truly equivariant to translations~\cite{Chaman_2021_CVPR} on equivariance to other symmetries.

Interestingly, we also find results that suggest equivariant networks offer performance increases to datasets that do not contain the relevant transformations, suggesting that using GCNNs might offer benefits other than equivariance to certain symmetries.
Furthermore, we find that relaxing equivariant constraints can be beneficial for network performance. However, relaxing equivariant constraints also allows networks to become biased towards the distribution of transformations in the training data.

\section{Limitations and Future Work}

The symmetries our method applies to are limited to rotations and reflections. While these symmetries are relevant, future work can include other symmetries, or more rotations in the group dimension, such as all rotations of $45^{\circ}$, rather than rotations of $90^{\circ}$.

We found experimentally that padding has a large influence on how well a GCNN generalises to unseen rotations, similar to CNNs~\cite{osman_location}. We found no explanation, but believe it is worth further investigating.

Section~\ref{ssec:breaking_eq} suggests that equivariant layers are more desirable at some depths than others, since the equivariance error drops at some depths and rises at others.
An interesting future work would be making a robust analysis of desirability of equivariance at different depths in a network.

Finally, we welcome further investigations of our results on PatchCamelyon, where we found that the approximately equivariant network learned to break its equivariance to increase performance, even on a problem  supposedly  invariant to rotation. With the rise of relaxed equivariant constraints~\cite{romero_partial_eq, relaxing_eq}, an interesting question to ask would be whether we are actually achieving better performance or simply exploiting unknown biases in data or in the network.

\small
\smallskip\noindent\textbf{Acknowledgements.} 
This project is (partly) financed by the Dutch Research Council (NWO) (project VI.Vidi.192.100).

{\small
\bibliographystyle{ieee_fullname}
\bibliography{egbib}

\begin{thebibliography}{10}\itemsep=-1pt

\bibitem{c3po}
Piyush Bagad, Floor Eijkelboom, Mark Fokkema, Danilo de Goede, Paul Hilders,
  and Miltiadis Kofinas.
\newblock C-3{PO}: Towards rotation equivariant feature detection and
  description.
\newblock In {\em 3rd Visual Inductive Priors for Data-Efficient Deep Learning
  Workshop}, 2022.

\bibitem{bekkers_roto-translation}
Erik~J Bekkers, Maxime~W Lafarge, Mitko Veta, Koen A~J Eppenhof, Josien P~W
  Pluim, and Remco Duits.
\newblock Roto-{Translation} {Covariant} {Convolutional} {Networks} for
  {Medical} {Image} {Analysis}.
\newblock In Alejandro~F. Frangi, Julia~A. Schnabel, Christos Davatzikos,
  Carlos Alberola-López, and Gabor Fichtinger, editors, {\em Medical {Image}
  {Computing} and {Computer} {Assisted} {Intervention} – {MICCAI} 2018},
  pages 440--448. Springer International Publishing, 2018.

\bibitem{Chaman_2021_CVPR}
Anadi Chaman and Ivan Dokmanic.
\newblock Truly shift-invariant convolutional neural networks.
\newblock In {\em Proceedings of the IEEE/CVF Conference on Computer Vision and
  Pattern Recognition (CVPR)}, pages 3773--3783, June 2021.

\bibitem{simCLR}
Ting Chen, Simon Kornblith, Mohammad Norouzi, and Geoffrey Hinton.
\newblock A simple framework for contrastive learning of visual
  representations.
\newblock In {\em International conference on machine learning}, pages
  1597--1607. PMLR, 2020.

\bibitem{GCNN}
Taco Cohen and Max Welling.
\newblock Group equivariant convolutional networks.
\newblock In Maria~Florina Balcan and Kilian~Q. Weinberger, editors, {\em
  Proceedings of The 33rd International Conference on Machine Learning},
  volume~48 of {\em Proceedings of Machine Learning Research}, pages
  2990--2999, New York, New York, USA, 2016. PMLR.

\bibitem{manifold_1}
Taco~S. Cohen, Maurice Weiler, Berkay Kicanaoglu, and Max Welling.
\newblock Gauge equivariant convolutional networks and the icosahedral cnn.
\newblock In {\em International Conference on Machine Learning}, 2019.

\bibitem{manifold_2}
Pim De~Haan, Maurice Weiler, Taco Cohen, and Max Welling.
\newblock Gauge equivariant mesh cnns: Anisotropic convolutions on geometric
  graphs.
\newblock {\em arXiv preprint arXiv:2003.05425}, 2020.

\bibitem{Lie_equivariance}
Nima Dehmamy, Robin Walters, Yanchen Liu, Dashun Wang, and Rose Yu.
\newblock Automatic symmetry discovery with lie algebra convolutional network.
\newblock In M. Ranzato, A. Beygelzimer, Y. Dauphin, P.S. Liang, and J.~Wortman
  Vaughan, editors, {\em Advances in Neural Information Processing Systems},
  volume~34, pages 2503--2515. Curran Associates, Inc., 2021.

\bibitem{imagenet}
Jia Deng, Wei Dong, Richard Socher, Li-Jia Li, Kai Li, and Li Fei-Fei.
\newblock Imagenet: A large-scale hierarchical image database.
\newblock In {\em 2009 IEEE Conference on Computer Vision and Pattern
  Recognition}, pages 248--255, 2009.

\bibitem{mnist}
Li Deng.
\newblock The mnist database of handwritten digit images for machine learning
  research.
\newblock {\em IEEE Signal Processing Magazine}, 29(6):141--142, 2012.

\bibitem{equimod}
Alexandre Devillers and Mathieu Lefort.
\newblock Equimod: An equivariance module to improve self-supervised learning.
\newblock {\em arXiv preprint arXiv:2211.01244}, 2022.

\bibitem{data_aug}
Jonas Geiping, Micah Goldblum, Gowthami Somepalli, Ravid Shwartz-Ziv, Tom
  Goldstein, and Andrew~Gordon Wilson.
\newblock How much data are augmentations worth? an investigation into scaling
  laws, invariance, and implicit regularization.
\newblock {\em arXiv preprint arXiv:2210.06441}, 2022.

\bibitem{rot_eq_sat}
Jiaming Han, Jian Ding, Nan Xue, and Gui-Song Xia.
\newblock Redet: A rotation-equivariant detector for aerial object detection.
\newblock In {\em Proceedings of the IEEE/CVF Conference on Computer Vision and
  Pattern Recognition (CVPR)}, pages 2786--2795, 2021.

\bibitem{resnets}
Kaiming He, Xiangyu Zhang, Shaoqing Ren, and Jian Sun.
\newblock Deep residual learning for image recognition.
\newblock In {\em Proceedings of the IEEE conference on computer vision and
  pattern recognition}, pages 770--778, 2016.

\bibitem{aprox_trans_1}
Md~Amirul Islam, Matthew Kowal, Sen Jia, Konstantinos~G Derpanis, and Neil~DB
  Bruce.
\newblock Global pooling, more than meets the eye: Position information is
  encoded channel-wise in cnns.
\newblock In {\em Proceedings of the IEEE/CVF International Conference on
  Computer Vision}, pages 793--801, 2021.

\bibitem{separate_2}
Max Jaderberg, Karen Simonyan, Andrew Zisserman, et~al.
\newblock Spatial transformer networks.
\newblock {\em Advances in neural information processing systems}, 28, 2015.

\bibitem{pooling}
Dong-Hwan Jang, Sanghyeok Chu, Joonhyuk Kim, and Bohyung Han.
\newblock Pooling revisited: Your receptive field is suboptimal.
\newblock In {\em Proceedings of the IEEE/CVF Conference on Computer Vision and
  Pattern Recognition}, pages 549--558, 2022.

\bibitem{symmetry_nature}
Iain~G. Johnston, Kamaludin Dingle, Sam~F. Greenbury, Chico~Q. Camargo,
  Jonathan P.~K. Doye, Sebastian~E. Ahnert, and Ard~A. Louis.
\newblock Symmetry and simplicity spontaneously emerge from the algorithmic
  nature of evolution.
\newblock {\em Proceedings of the National Academy of Sciences},
  119(11):e2113883119, 2022.

\bibitem{osman_location}
Osman~Semih Kayhan and Jan C~van Gemert.
\newblock On translation invariance in cnns: Convolutional layers can exploit
  absolute spatial location.
\newblock In {\em Proceedings of the IEEE/CVF Conference on Computer Vision and
  Pattern Recognition}, pages 14274--14285, 2020.

\bibitem{adam}
Diederik~P. Kingma and Jimmy Ba.
\newblock Adam: A method for stochastic optimization.
\newblock {\em CoRR}, abs/1412.6980, 2014.

\bibitem{cifar}
Alex Krizhevsky, Geoffrey Hinton, et~al.
\newblock Learning multiple layers of features from tiny images.
\newblock 2009.

\bibitem{cnns_translation}
Alex Krizhevsky, Ilya Sutskever, and Geoffrey Hinton.
\newblock Imagenet classification with deep convolutional neural networks.
\newblock {\em Neural Information Processing Systems}, 25, 2012.

\bibitem{separate_3}
Denis Kuzminykh, Daniil Polykovskiy, and Alexander Zhebrak.
\newblock Extracting invariant features from images using an equivariant
  autoencoder.
\newblock In Jun Zhu and Ichiro Takeuchi, editors, {\em Proceedings of The 10th
  Asian Conference on Machine Learning}, volume~95 of {\em Proceedings of
  Machine Learning Research}, pages 438--453. PMLR, 2018.

\bibitem{roto-trans_hisp}
Maxime~W Lafarge, Erik~J Bekkers, Josien~PW Pluim, Remco Duits, and Mitko Veta.
\newblock Roto-translation equivariant convolutional networks: Application to
  histopathology image analysis.
\newblock {\em Medical Image Analysis}, 68:101849, 2021.

\bibitem{deforestation}
Joshua Mitton and Roderick Murray-Smith.
\newblock Rotation equivariant deforestation segmentation and driver
  classification.
\newblock {\em arXiv preprint arXiv:2110.13097}, 2021.

\bibitem{RIC-CNN}
Hanlin Mo and Guoying Zhao.
\newblock Ric-cnn: Rotation-invariant coordinate convolutional neural network.
\newblock {\em arXiv preprint arXiv:2211.11812}, 2022.

\bibitem{flowers102}
Maria-Elena Nilsback and Andrew Zisserman.
\newblock Automated flower classification over a large number of classes.
\newblock In {\em Indian Conference on Computer Vision, Graphics and Image
  Processing}, 2008.

\bibitem{natural_equivariance}
Chris Olah, Nick Cammarata, Chelsea Voss, Ludwig Schubert, and Gabriel Goh.
\newblock Naturally occurring equivariance in neural networks.
\newblock {\em Distill}, 5, 2020.

\bibitem{CNN}
Keiron O'Shea and Ryan Nash.
\newblock An introduction to convolutional neural networks.
\newblock {\em arXiv preprint arXiv:1511.08458}, 2015.

\bibitem{romero2020attentive}
David~W Romero, Erik~J Bekkers, Jakub~M Tomczak, and Mark Hoogendoorn.
\newblock Attentive group equivariant convolutional networks.
\newblock {\em arXiv preprint arXiv:2002.03830}, 2020.

\bibitem{romero_partial_eq}
David~W Romero and Suhas Lohit.
\newblock Learning equivariances and partial equivariances from data.
\newblock {\em arXiv preprint arXiv:2110.10211}, 2021.

\bibitem{rotation_transformer}
Kai~Sheng Tai, Peter Bailis, and Gregory Valiant.
\newblock Equivariant transformer networks.
\newblock In {\em International Conference on Machine Learning}, pages
  6086--6095. PMLR, 2019.

\bibitem{relaxing_eq}
Tycho~FA van~der Ouderaa, David~W Romero, and Mark van~der Wilk.
\newblock Relaxing equivariance constraints with non-stationary continuous
  filters.
\newblock {\em arXiv preprint arXiv:2204.07178}, 2022.

\bibitem{marginal_1}
Tycho~F.A. van~der Ouderaa and Mark van~der Wilk.
\newblock Learning invariant weights in neural networks.
\newblock In James Cussens and Kun Zhang, editors, {\em Proceedings of the
  Thirty-Eighth Conference on Uncertainty in Artificial Intelligence}, volume
  180 of {\em Proceedings of Machine Learning Research}, pages 1992--2001.
  PMLR, 2022.

\bibitem{marginal_2}
Mark van~der Wilk, Matthias Bauer, ST John, and James Hensman.
\newblock Learning invariances using the marginal likelihood.
\newblock {\em Advances in Neural Information Processing Systems}, 31, 2018.

\bibitem{PCAM_rot_eq}
Bastiaan~S Veeling, Jasper Linmans, Jim Winkens, Taco Cohen, and Max Welling.
\newblock Rotation equivariant cnns for digital pathology.
\newblock In {\em Medical Image Computing and Computer Assisted
  Intervention--MICCAI 2018: 21st International Conference, Granada, Spain,
  September 16-20, 2018, Proceedings, Part II 11}, pages 210--218. Springer,
  2018.

\bibitem{manifold_3}
Maurice Weiler, Patrick Forr{\'e}, Erik Verlinde, and Max Welling.
\newblock Coordinate independent convolutional networks--isometry and gauge
  equivariant convolutions on riemannian manifolds.
\newblock {\em arXiv preprint arXiv:2106.06020}, 2021.

\bibitem{group_equivariant_subsampling}
Jin Xu, Hyunjik Kim, Thomas Rainforth, and Yee Teh.
\newblock Group equivariant subsampling.
\newblock {\em Advances in Neural Information Processing Systems},
  34:5934--5946, 2021.

\bibitem{blurpool}
Richard Zhang.
\newblock Making convolutional networks shift-invariant again.
\newblock In {\em International conference on machine learning}, pages
  7324--7334. PMLR, 2019.

\bibitem{separate_1}
Allan Zhou, Tom Knowles, and Chelsea Finn.
\newblock Meta-learning symmetries by reparameterization.
\newblock {\em arXiv preprint arXiv:2007.02933}, 2020.

\end{thebibliography}
}

\end{document}